\documentclass[runningheads]{llncs}
\usepackage{hyperref}
\usepackage{amssymb}
\usepackage{tabularx}
\usepackage[utf8]{inputenc}
\usepackage{CJKutf8}
\usepackage{comment}
\newcommand{\zh}[1]{\begin{CJK*}{UTF8}{gbsn}#1\end{CJK*}}
\usepackage{graphicx}
\usepackage{multirow}
\usepackage{xcolor}

\begin{document}
\title{Puzzle Pieces Picker: Deciphering Ancient Chinese Characters with Radical Reconstruction}

\titlerunning{Puzzle Pieces Picker}
%
\author{Pengjie Wang\inst{1} \and
Kaile Zhang\inst{1} \and
Xinyu Wang\inst{2}\thanks{Corresponding author. Email: xinyu.wang02@adelaide.edu.au} \and
Shengwei Han\inst{3} \and Yongge Liu\inst{3}\and Lianwen Jin\inst{4}\and Xiang Bai\inst{1} \and Yuliang Liu\inst{1} }
\authorrunning{P. Wang et al.}
%
\institute{Huazhong University of Science and Technology, Wuhan, 430074, China \and The University of Adelaide, SA, 5005, Australia
\and Anyang Normal University, Anyang, 450000, China
\and South China University of Technology, Guangzhou, 510641, China}

\maketitle              
\vspace{-0.2cm}
\begin{abstract}
Oracle Bone Inscriptions is one of the oldest existing forms of writing in the world. However, due to the great antiquity of the era, a large number of Oracle Bone Inscriptions (OBI) remain undeciphered, making it one of the global challenges in the field of paleography today. This paper introduces a novel approach, namely Puzzle Pieces Picker (P$^3$), to decipher these enigmatic characters through radical reconstruction. We deconstruct OBI into foundational strokes and radicals, then employ a Transformer model to reconstruct them into their modern counterparts, offering a groundbreaking solution to ancient script analysis. To further this endeavor, a new Ancient Chinese Character Puzzles (ACCP) dataset was developed, comprising an extensive collection of character images from seven key historical stages, annotated with detailed radical sequences. The experiments have showcased considerable promising insights, underscoring the potential and effectiveness of our approach in deciphering the intricacies of ancient Chinese scripts. Through this novel dataset and methodology, we aim to bridge the gap between traditional philology and modern document analysis techniques, offering new insights into the rich history of Chinese linguistic heritage. 

\end{abstract}

\section{Introduction}

\begin{figure}[t!]
    \centering
    \includegraphics[width=\linewidth]{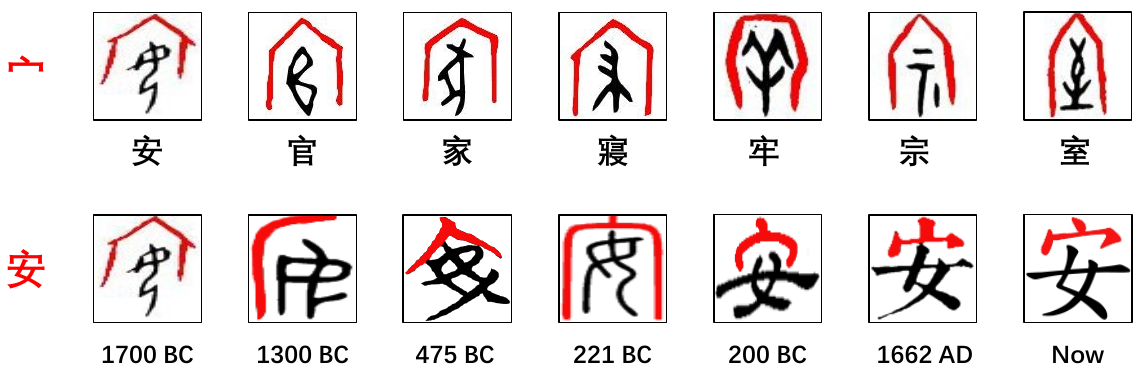}
    \caption{The consistency and evolution of the roof-like component ``\zh{宀}" in Chinese characters. The first row demonstrates the uniformity in the representation of the radical among various characters during the Oracle Bone Inscriptions period. The second row depicts the evolutionary path of the character ``\zh{安}", highlighting the changes in the roof-like component from 1700 BC through subsequent historical stages to the standardized form in contemporary Chinese script.}
    \label{fig:intro}
    \vspace{-0.5cm}
\end{figure}

Chinese characters represent one of the few writing systems in the world with a documented history of continuous evolution. Originating from the Oracle Bone Inscriptions (OBI) of the Shang Dynasty~\cite{boltz1986early}, they have evolved over more than 3,000 years into modern standard Chinese characters, undergoing significant changes in both quantity and form. As a result, many ancient characters have lost their meanings, making it difficult to find their modern equivalents and interpretations. These ancient scripts, particularly the OBI, contain valuable information about the human and geographic landscape of millennia ago, offering crucial insights for archaeology and history, thus attracting the interest of many scholars. However, more than two-thirds of the over 4,500 discovered OBI characters still have unknown meanings, presenting a significant challenge to interpreting complete oracle bone documents~\cite{bazerman2009handbook}.

In recent years, with the advancement of Artificial Intelligence (AI), particularly in document analysis and natural language processing, scholars have begun to digitize ancient characters~\cite{suzuki2013survey} and employ machine learning models to recognize them. However, these approaches often rely on existing Optical Character Recognition (OCR) or object detection algorithms and depend on training data with known labels, which limits their ability to generalize to texts with unknown labels, thus hindering their capacity to decipher ancient texts with unknown meanings. Pioneering approaches have also made initial attempts to decipher unknown ancient Chinese scripts using AI algorithms. For instance, Diao et al.~\cite{diao2023toward} and Lin et al.~\cite{lin2022radical} have explored the potential of training models on manually annotated radicals in OBIs, enabling these models to recognize and infer meanings for characters they have never seen before. This method's advantage lies in its utilization of the models' generalization abilities, extending beyond the constraints of training solely on data with known labels to interpret previously unlabeled characters. While these innovative strategies have shown promise, they are not without their challenges. The intensive labor required for manual annotation restricts these studies to relatively small datasets, often limited to a few hundred characters. Additionally, integrating cross-era character evolution to enhance deciphering efforts poses another complex challenge.

In this paper, inspired by the concept of a jigsaw puzzle, we innovatively propose a method for deciphering ancient Chinese characters named \textbf{P}uzzle \textbf{P}ieces \textbf{P}icker (P$^3$). This approach is fundamentally grounded in two key observations: the \textit{shape consistency} in the forms of radicals from the same era, and the \textit{structured consistency} across different epochs in their evolutionary trajectory. Specifically, as illustrated in Figure~\ref{fig:intro}, the first row displays various OBI composed of the roof-like component ``\zh{宀}", showcasing a remarkable uniformity in their written forms. The second row delineates the evolutionary journey of the Chinese character ``\zh{安}" from 1700 BC to its contemporary form. It is evident that despite the variations in the portrayal of the ``\zh{宀}" radical over time, the structural configuration of the character has remained consistent, maintaining an upper-lower structure where the ``\zh{宀}" is positioned above the ``\zh{女}" character.

Building upon the above findings, and taking into account that Chinese characters from different periods are composed of similar elements such as radicals, it becomes feasible to deduce the manifestation of a radical in other periods once its meaning in a specific era is understood. Consequently, as depicted in Figure~\ref{fig:puzzle}, we treat the decipherment of ancient Chinese texts as a puzzle game, wherein ancient characters from various periods are deconstructed into distinct puzzle pieces according to their radicals and other components, with the original textual structure sequences as recipes for reassembly. Subsequently, our proposed P$^3$ framework employs a Transformer-based~\cite{vaswani2017attention} sequence prediction model trained to examine the evolutionary patterns of components across different epochs. Thus, during the inference phase, when introduced to new, previously unseen ancient texts, it is capable of selecting the appropriate puzzle pieces and providing the reassembly recipe, thereby unveiling the decipherment results. Specifically, the main contributions of this paper are as follows:

\begin{figure}[t!]
    \centering
    \includegraphics[width=\linewidth]{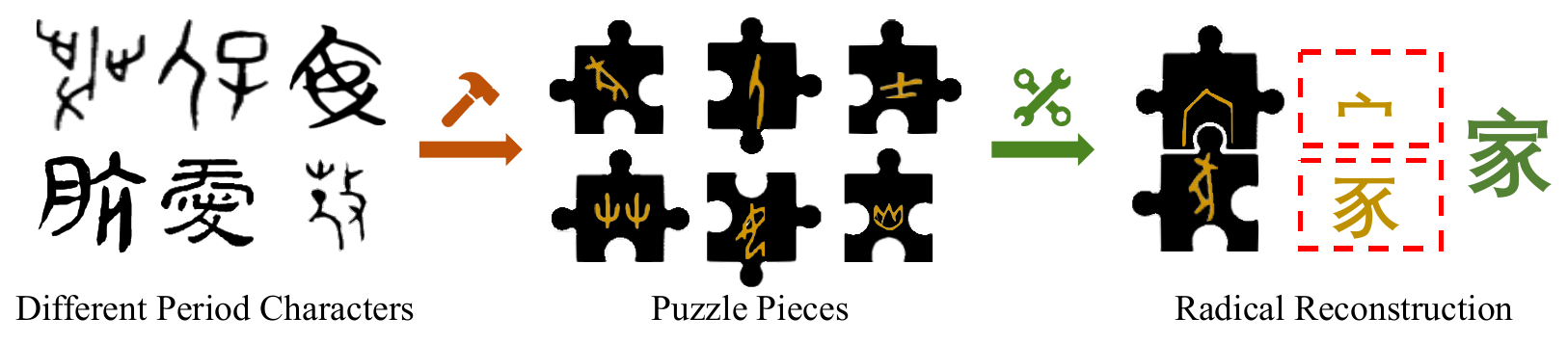}
    \caption{The decipherment of ancient Chinese characters is treated as a puzzle-solving game, where characters from various periods are first broken down into pieces according to radical strokes. These pieces are then analyzed by the proposed P$^3$ that examines potential evolutionary patterns. In the inference stage, the model predicts a reconstruction recipe when presented with new, unseen samples, thus aiding in decipherment.}
    \label{fig:puzzle}
    \vspace{-0.4cm}
\end{figure}

\begin{itemize}

\item We have constructed a large-scale dataset named Ancient Chinese Character Puzzles (ACCP), comprising nearly 90,000 categories of Chinese characters from 7 distinct historical periods, totaling over 340,000 images. For each character, we provide detailed annotations, including its category and radical sequence.

\item We introduce a novel approach, Puzzle Pieces Picker (P$^3$), a sequence prediction model based on a Transformer architecture. To the best of our knowledge, this represents the first attempt to decipher ancient texts through a method of deconstructing and reconstructing characters from different eras.

\item We conducted extensive experiments to evaluate the performance of the proposed P$^3$ model in deciphering texts from various periods and assess the impact of training on cross-era data. The results demonstrate that P$^3$ shows promising effectiveness in simulated deciphering tasks, highlighting its potential in advancing the field of ancient language analysis.

\end{itemize}

\section{Related Work}

The incorporation of AI, particularly OCR and other document analysis techniques, into the study of ancient Chinese characters, marks a significant shift in the way historical documents are analyzed. 

 \textbf{Ancient text detection and recognition.} The challenge lies in accurately detecting~\cite{meng2019oracle,xing2019oracle,2020Kmeans,2021FasterR-CNN,2021NMS,fang2022automatic} and recognizing~\cite{2016Sketch,2019KNN,han2020self,lin2022radical,2022wang} ancient characters from scanned images of historical documents. This task shares considerable similarities with OCR and general object detection, leading much of the existing research to focus on adapting established models to meet the unique challenges of working with ancient languages. For instance, Meng et al.~\cite{meng2019oracle}, Liu et al.~\cite{2021FasterR-CNN}, and Yang et al.~\cite{2021FasterR-CNN} have applied well-known general object detectors like SSD~\cite{liu2016ssd}, Faster R-CNN~\cite{girshick2015fast}, and YOLO-v4~\cite{bochkovskiy2020yolov4} to the detection of ancient Chinese characters. Their efforts were primarily aimed at tackling the intricate challenges posed by the dense text arrangement in ancient manuscripts. 

In the realm of recognition, OCR technology has already achieved remarkable success in processing modern documents and scene texts. However, the challenges associated with ancient languages are notably more complex due to the limited availability of training data, which is often further complicated by issues such as missing pieces, damage, and noise in historical documents. For example, datasets such as OBI-100~\cite{OBI100}, OBI-125~\cite{OBI125}, Oracle-20k~\cite{20K}, and OBC306~\cite{OBC306} each contain only a few hundred Oracle Bone Inscription characters, and some of these characters are affected by significant noise. Consequently, research efforts have increasingly shifted towards the creation of innovative data augmentation strategies and the implementation of few-shot learning techniques. For example, Han et al.~\cite{han2020self} introduced the ORC-BERT Augmentor, a novel data augmentation approach pre-trained through self-supervised learning, specifically designed for the recognition of ancient characters with limited training data. Lin et al.~\cite{lin2022radical} developed a radical extraction and recognition framework, which approaches ancient Chinese characters as combinations of radicals and identifies them by detecting these constituent radicals. Wang et al.~\cite{2022wang} proposed the structure-texture separation network, which serves as an end-to-end learning framework designed to perform disentanglement, transformation, adaption, and recognition tasks on ancient characters in a unified process.

\textbf{Zero-shot and few-shot learning of Chinese characters.} Despite these methods' success in employing CV for ancient text detection and recognition, their reliance on data from known categories limits their applicability to deciphering tasks, which require testing on unlabeled data. In response, some studies~\cite{diao2023toward,zhang2018radical,wang2018denseran,2021RCN,CAO2020107488,chang2022sundial,2021Zhang} have turned to zero-shot or few-shot learning, aiming to apply methods trained on known data to new, unseen texts, thus facilitate the decipherment of unknown characters. For modern Chinese characters, some research~\cite{zhang2018radical,wang2018denseran,2021RCN,CAO2020107488} has concentrated on the analysis of strokes and radical structures within modern Chinese characters to facilitate their interpretation. Li et al.~\cite{2021RCN}, proposed the Radical Counter Network, designed to detect and quantify radicals and their spatial configurations within characters. Similarly, Cao et al.~\cite{CAO2020107488} delved into the hierarchical decomposition of modern Chinese characters, devising an algorithm that translates characters into vectors representative of their structural elements. However, due to the significant differences between ancient and modern Chinese characters in terms of radicals and structures, these previous methods cannot be directly applied to decryption tasks of ancient Chinese characters. For ancient Chinese characters, Chang et al.~\cite{chang2022sundial} developed a four-stage cascade GAN, trained using images of ancient Chinese scripts from four distinct historical epochs. This innovative model can simulate the evolutionary trajectory of specific characters, allowing for the generation of potential decipherment clues when presented with images of unknown texts. Diao et al.~\cite{diao2023toward} conducted zero-shot learning on Oracle Bone Inscriptions by manually annotating radicals, employing common object detection methods. However, these prior approaches still exhibit shortcomings, including excessive dependence on manually annotated radical information and a lack of validation on large-scale datasets.

By contrast, the proposed model can extract and learn from the evolutionary patterns of radicals in nearly 90,000 Chinese characters across seven periods. It can automatically decompose more radical pieces for the puzzle, increasing the likelihood of reconstructing the ancient characters that need to be deciphered.
\vspace{-0.2cm}

\section{The ACCP Dataset}
\label{sec:dataset}

\begin{figure}[t!]
    \centering
    \includegraphics[width=0.8\linewidth]{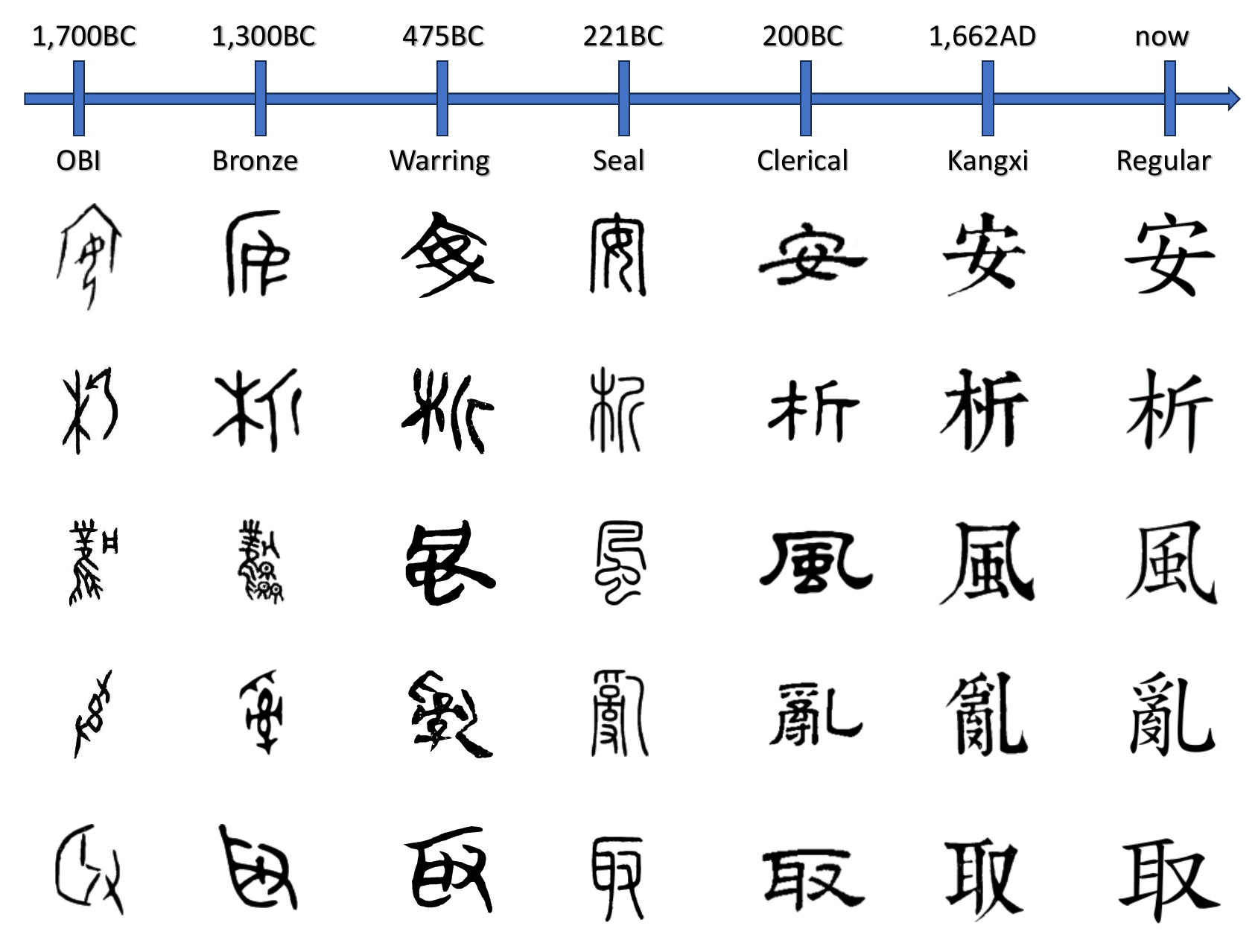}
    \caption{Examples of the evolution of Chinese characters across seven historical periods in our ACCP dataset. Each row showcases a category of characters, while each column corresponds to a specific period, illustrating the developmental trajectory within the same character category.}
    \label{fig:Dataset}
    \vspace{-0.4cm}
\end{figure}

Chinese characters have undergone a millennia-long evolution, with their iconic forms broadly categorized into seven historical stages: Oracle Bone Inscriptions, Bronze Inscriptions, Warring States Script, Seal Script, Clerical Script, the script of the Kangxi Dictionary, and Modern Regular Chinese Characters. Despite the vast transformation from the original OBI to contemporary characters, there are notable similarities between scripts from successive periods. For example, as illustrated in Figure~\ref{fig:Dataset}, the temporal gap between OBI and Bronze Inscriptions is merely about 300 years, leading to substantial similarities in their forms. Following the unification of the six states by the Qin dynasty in 200 BC, Clerical Script emerged as the predominant script, closely resembling modern Chinese characters. This clear evolutionary trajectory facilitates the cross-validation of character forms across different eras, a common approach in the study of ancient Chinese scripts. However, existing datasets rarely provide a comprehensive collection of characters across these varied historical stages. Among the larger datasets, EVOBC stands out, yet it only encompasses characters from four distinct periods: Bronze Inscriptions, Warring States Script, Seal Script, and Clerical Script. To better leverage the evolutionary characteristics of Chinese characters from deciphering ancient scripts, we expanded upon one of the largest OBI datasets, HUST-OBC~\cite{wang2024open}, and the evolutionary dataset EVOBC~\cite{guan2024open}. As demonstrated in Table~\ref{tab:Dataset}, our extension incorporates additional data from the Kangxi Dictionary, a comprehensive work from the Qing Dynasty known for its detailed compilation of character forms, and modern Regular Characters. This expanded collection forms the basis of our dataset, named Ancient Chinese Character Puzzles \textbf{(ACCP)}, designed specifically to advance the study and decipherment of ancient Chinese texts. ACCP will be available at \url{https://github.com/Pengjie-W/Puzzle-Pieces-Picker}

\begin{table}[t!!]
\begin{tabular}{cc|c|c|c}
\hline
\multicolumn{1}{c|}{Source}                     & Script Type              & Time Period      & \#Category & \#Sample  \\ \hline
\multicolumn{1}{c|}{HUST-OBC~\cite{wang2024open}}                   & Oracle Bone Inscriptions & 1,700BC - 1100BC & 1,781    & 77,064  \\ \hline
\multicolumn{1}{c|}{\multirow{4}{*}{EVOBC~\cite{guan2024open}}}     & Bronze Inscription       & 1,300BC - 200BC  & 4,801    & 42,573  \\
\multicolumn{1}{c|}{}                           & Warring States script    & 475BC - 211BC    & 5,343    & 80,119  \\
\multicolumn{1}{c|}{}                           & Seal Script              & 221BC - 420AD    & 14,158   & 29,751  \\
\multicolumn{1}{c|}{}                           & Clerical Script          & 200BC - 220AD    & 2,890    & 3,568   \\ \hline
\multicolumn{1}{c|}{\multirow{2}{*}{Extension}} & Kangxi Dictionary        & 1,662AD - 1722AD & 9,354    & 9,354   \\
\multicolumn{1}{c|}{}                           & Regular Character & Now              & 88,899   & 103,915 \\ \hline
\multicolumn{2}{c|}{Total}                                                 & 1,700BC - Now    & 88,901   & 346,344 \\ \hline
\end{tabular}
\caption{The ACCP dataset spans over 3,000 years, covering seven script types from various historical periods, and comprises nearly 90k categories with more than 340k image samples.}
\label{tab:Dataset}
\vspace{-0.8cm}
\end{table}
\vspace{-0.4cm}
\section{Method}
\vspace{-0.2cm}

In the introduction, we briefly outlined the concept of the Puzzle Piece Picker (P$^3$) model, which is inspired by jigsaw puzzles and primarily involves two main steps: \textit{deconstruction} and \textit{reconstruction}. In this section, we will delve into the specific implementation details of the P$^3$ model.
\vspace{-0.2cm}
\subsection{Radical Decomposition}

The P$^3$ model facilitates the deciphering of unknown ancient Chinese characters by deconstructing them according to radicals and then reconstructing them based on the rules of radicals in modern Chinese characters. This process necessitates the use of training data annotated at the radical level, which aids the model in learning the local structures of radical components. However, as outlined in Section~\ref{sec:dataset}, the ACCP dataset, which is expanded from HUST-OBC~\cite{wang2024open} and EVOBC~\cite{guan2024open}, lacks such annotations. Manually annotating using image editing software like Photoshop is prohibitively expensive due to the high costs involved.

\begin{figure}[t!]
    \centering
    \includegraphics[width=\linewidth]{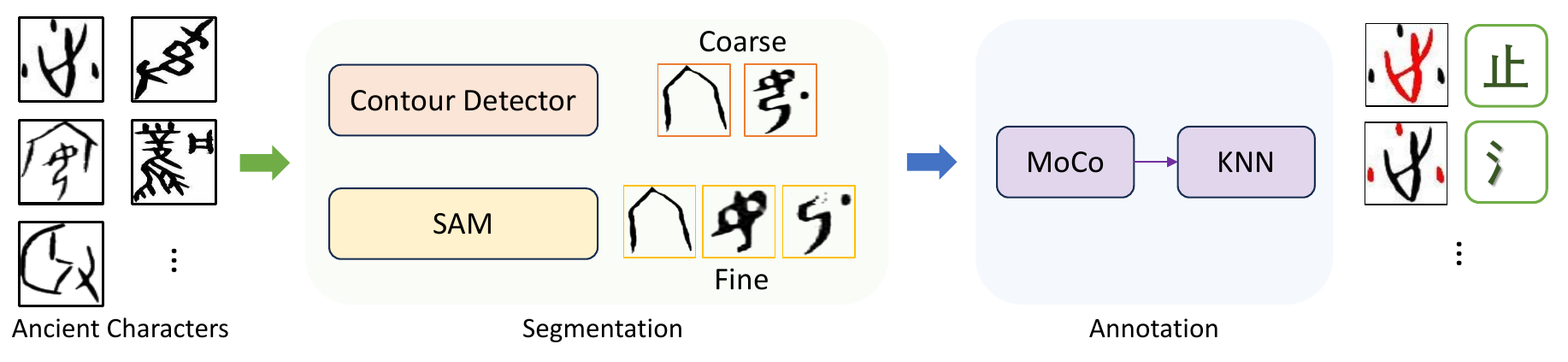}
    \caption{
Flowchart of radical deconstruction.}
    \label{fig:Generation}
    \vspace{-0.5cm}
\end{figure}

\noindent\textbf{Segmentation.} To acquire annotations at the radical level, a straightforward approach involves utilizing segmentation models to obtain segmentation masks for radicals. However, there isn't an off-the-shelf model capable of performing this task independently. Consequently, we devised a radical decomposition process that operates on both coarse and fine levels, as illustrated in Figure~\ref{fig:Generation}. Specifically, we employed a contour detector for coarse-grained segmentation and the Segment Anything Model (SAM)~\cite{kirillov2023segment}, a large-scale, general-purpose semantic segmentation model, for fine-grained segmentation.

\textit{Contour-based}: For contour-based decomposition, we developed a heuristic method that leverages OpenCV to detect the edges of ancient Chinese character images. By setting a distance threshold, we merged components within a certain range to delineate the radical components. This method is simple yet effective for characters with dispersed structures. However, its limitations become apparent with characters that are closely adjacent or overlapping, where it struggles to accurately separate individual elements.

\textit{SAM-based}: Regarding the SAM-based approach, SAM, has been pre-trained on a vast dataset and is adept at zero-shot generalization for segmenting virtually any object. Thus we also employed it to obtain radical masks for ancient character images. Compared to the contour-based method, SAM is more proficient at distinguishing radicals that are stuck together. However, its drawback lies in over-segmenting, where it might divide a single component into multiple fragments.

\begin{figure}[t!]
    \centering
    \includegraphics[width=0.6\linewidth]{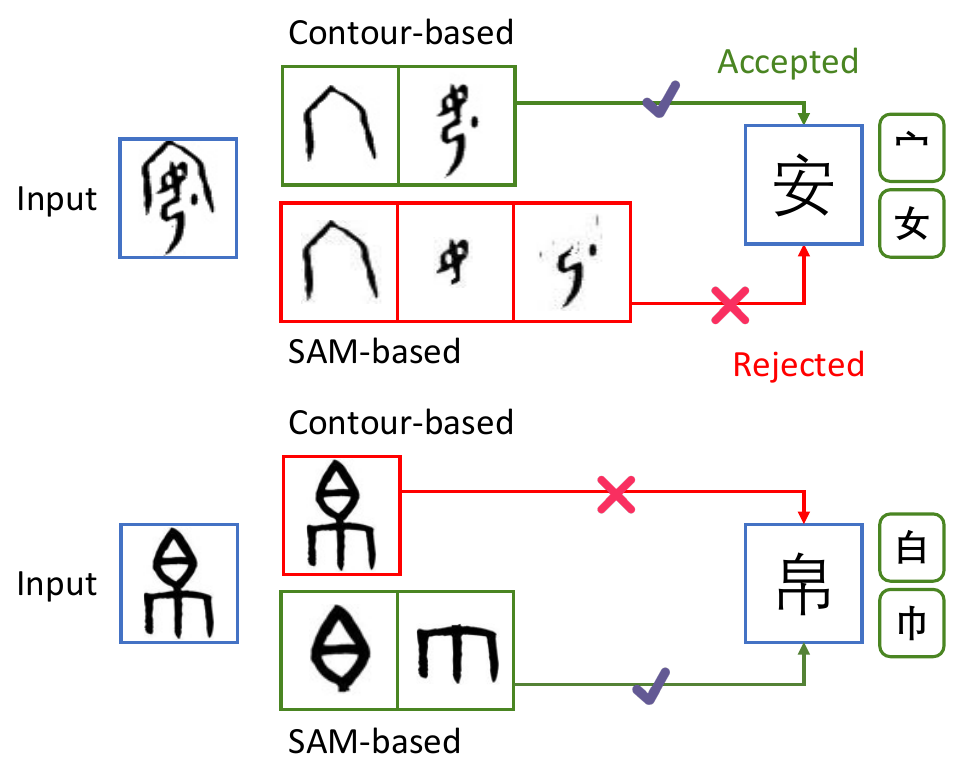}
    \caption{Contour-based and SAM-based methods yield different sets of masks for radicals. These segmented components are tested for their ability to reconstruct the specified modern Chinese characters. Successful reconstructions are retained as final annotations, while failed attempts indicate which segmentation masks to discard.}
    \label{fig:seg}
    \vspace{-0.6cm}
\end{figure}

As a result, each ancient character image yields two distinct sets of segmentation outcomes, one from the contour-based method and the other from the SAM-based approach, as depicted in Figure~\ref{fig:seg}. The effectiveness of each method can vary; for some characters, the SAM-based results may prove superior, while for others, the contour-based might be more effective. 

\noindent\textbf{Annotation.} Merely possessing segmentation masks is insufficient. It is also essential to acquire labels corresponding to the radicals of modern Chinese characters associated with these segmented components. To address this, we employed MoCo~\cite{he2020momentum} to learn the representations of different components and then grouped them using KNN. 

Specifically, \textit{MoCo}~\cite{he2020momentum} takes all of the radical pieces segmented by both contour-based and SAM-based methods as training samples, and maps them into a feature space, facilitating the measurement of their similarities. During each iteration, an image of a radical serves as the positive sample, while others are treated as negative samples. This setup aims to minimize the contrastive loss defined by the following equation:
\vspace{-0.1cm}
\begin{equation} 
\mathcal{L}_q=-\log \frac{\exp \left(q \cdot k_{+} / \tau\right)}{\sum_{i=1}^K \exp \left(q \cdot k_i / \tau\right)}
\vspace{-0.1cm}
\end{equation}

\noindent where $\tau$ represents the temperature parameter, which aids in controlling the separation of distribution, $K$ denotes the number of selected samples for comparison, $q$ is the feature vector derived from the enhanced positive sample, and $k_{+}$ is the feature vector of the actual positive sample. By iteratively optimizing this loss, our model effectively discriminates between the positive samples and a multitude of negative samples, thereby learning robust representations of radicals in ancient Chinese characters.

\begin{figure}[t!]
    \centering
    \includegraphics[width=0.8\linewidth]{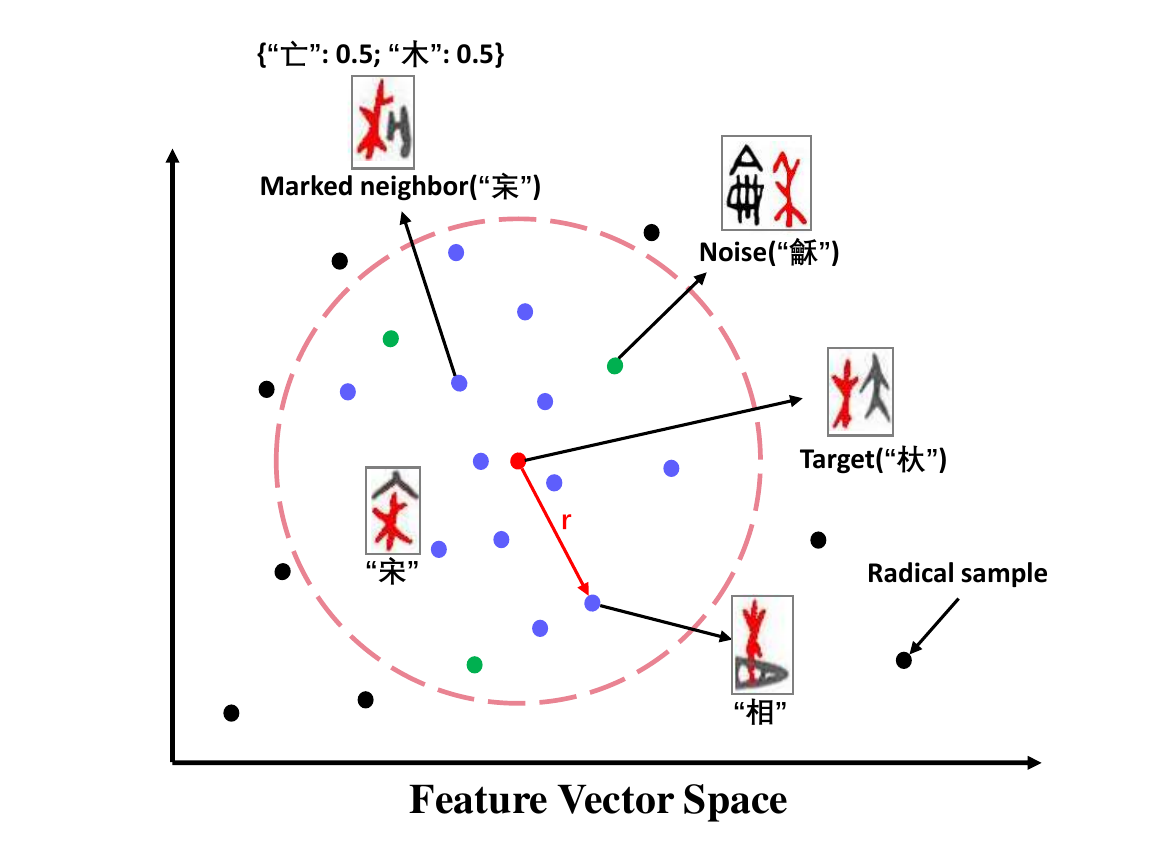}
    \caption{Illustration of the KNN clustering in the feature space. The target radical component is marked in red, surrounded by its 15 nearest neighbors within the delineated boundary. Green points represent similar, yet non-matching radicals, while blue points correspond to radicals sharing the same radical as the target, aiding in the accurate labeling process.}
    \label{fig:cluster}
    \vspace{-0.7cm}
\end{figure}

Following the completion of the training process, we utilized ResNet-50~\cite{he2016deep} to transform images of each radical into a 128-dimensional feature vector, denoted as ${f} \in {\mathbb{R}^{128}}$. This encoding facilitates the subsequent application of a \textit{KNN} algorithm for clustering and annotating these radicals. As illustrated in Figure~\ref{fig:cluster}, the procedure for an unlabeled radical component begins with identifying its $K$ nearest previously annotated neighbors within the feature space, utilizing Euclidean distances for this determination. These annotated neighbors are either components that have already been labeled or are part of an ancient Chinese character whose interpretation has been established. Each annotated component is associated with a dictionary where the dictionary keys represent potential radicals, and the values indicate confidence levels. This dictionary is initially populated based on the Ideographic Description Sequence (IDS) of the corresponding Chinese character, with keys derived from modern Chinese radical components. The confidence level assigned to each key is set to $\frac{1}{n}$, where $n$, typically ranging from 2 to 3, is the total number of radicals comprising the Chinese character. Upon identifying the marked neighbors and their associated radical dictionaries, we proceeded to refine the dictionary corresponding to the target ancient character radical:
\vspace{-0.28cm}
\begin{equation} 
F_c=softmax(\sum_{i=1}^K r_i f_c)
\vspace{-0.28cm}
\end{equation}

In this equation, $F_c$ represents the updated confidence level, $r_i$ signifies the Euclidean distance between the target component and a marked neighbor within the feature space, and $f_c$ denotes the confidence levels associated with the marked neighbor, with a value of 0 assigned if the key is absent from the dictionary. The final step involves determining the label for the target ancient character radical by selecting the dictionary key that exhibits the highest value. Ultimately, as shown in Figure~\ref{fig:seg}, the labeled radical components are utilized in attempts to reconstruct corresponding modern Chinese characters. Successful reconstructions are accepted as final annotations, while failures lead to rejection. This process effectively filters out inferior segmentation masks produced during the initial coarse and fine segmentation phases.

\begin{figure}[t!]
    \centering
    \includegraphics[width=0.85\linewidth]{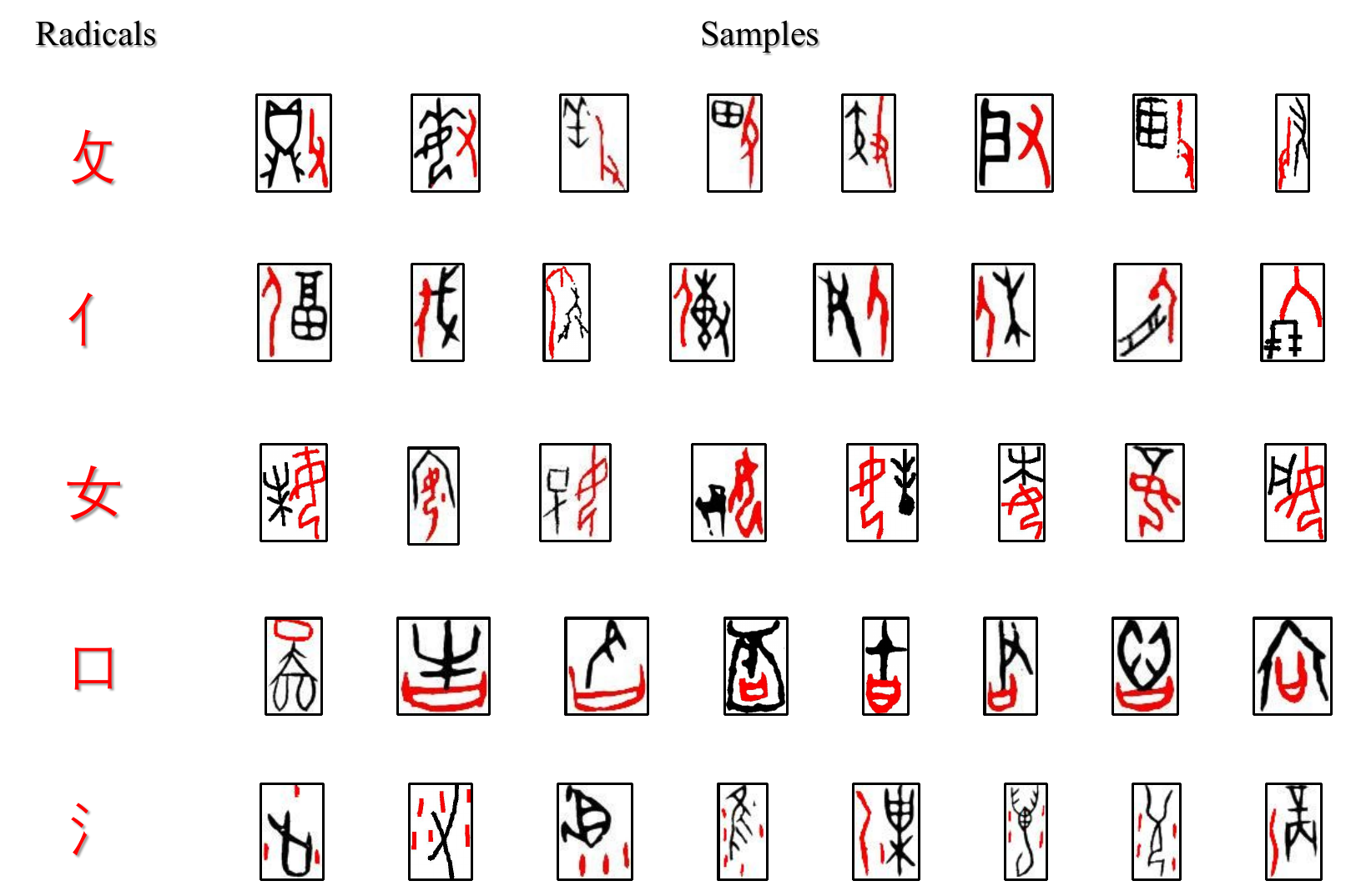}
    \caption{Examples of segmented radical components by radical decomposition.}
    \label{fig:results}
    \vspace{-0.5cm}
\end{figure}

Given that OBIs represent a historical phase with the most undeciphered characters and suffer from a limited number of training samples, we primarily employ radical decomposition techniques to synthesize supplementary radical data, thereby enriching the training dataset. This augmentation aids the model in examining the structures and shapes of radicals more effectively. It is important to note, however, that the proposed method for radical decomposition is adaptable not only to ancient Chinese characters from any period but also to other radical-based languages, such as Korean and Japanese. Figure~\ref{fig:results} showcases examples of radical-level annotations produced by our methods, demonstrating their effectiveness in segmenting ancient Chinese characters by radicals and in automatically assigning accurate labels.
\vspace{-0.4cm}
\subsection{Radical Reconstruction}
\vspace{-0.2cm}
\subsubsection{Preliminary.} 
\begin{figure}[t!]
    \centering
    \includegraphics[width=0.6\linewidth]{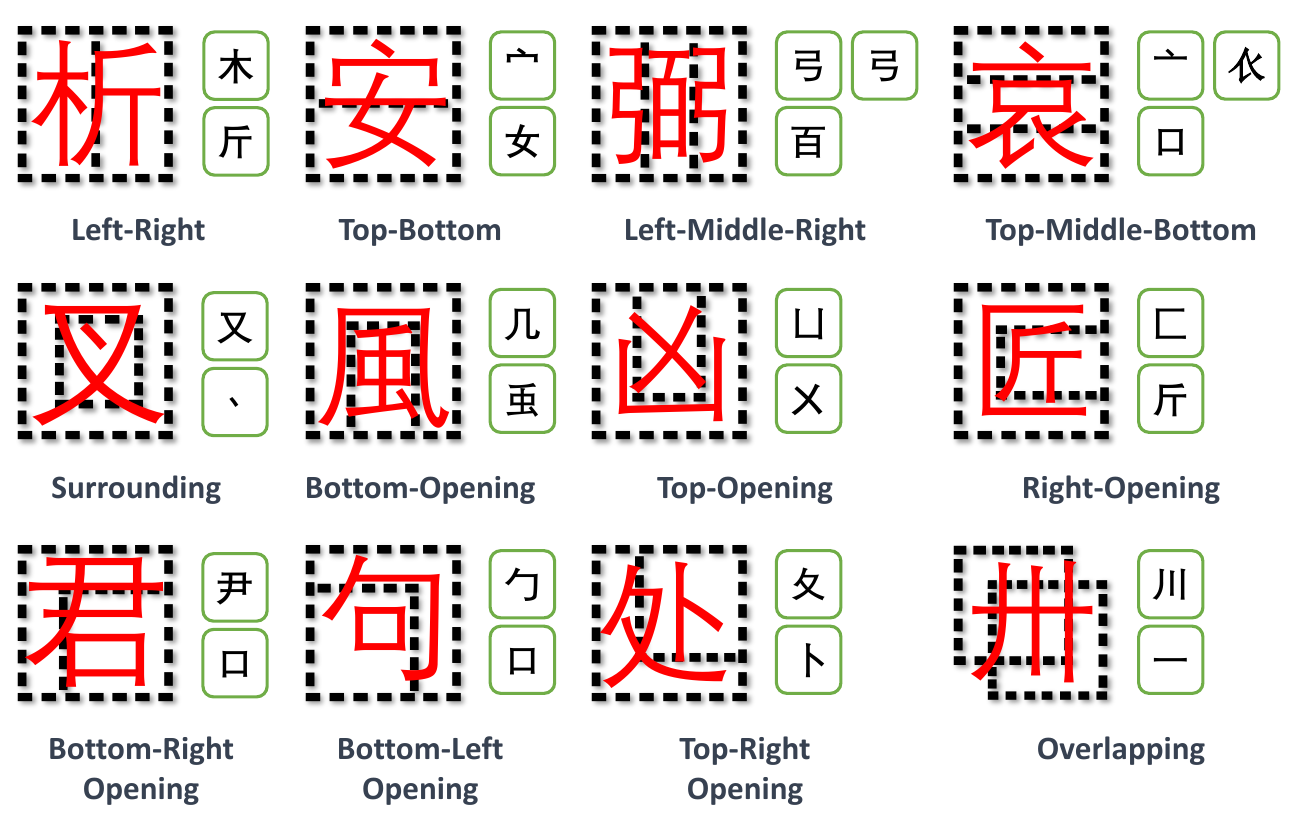}
    \caption{12 common structures of Chinese characters in IDS.}
    \label{fig:IDS}
    \vspace{-0.3cm}
\end{figure}

In the endeavor to decipher ancient Chinese characters, AI-assisted models play a pivotal role in suggesting possible modern equivalents for undeciphered characters. Our innovative P$^3$ model introduces a novel pipeline that first deconstructs ancient characters into their radical components and then reconstructs these components into modern Chinese characters. This reconstruction process is guided by an understanding of the structural principles of modern Chinese writing. As demonstrated in Figure~\ref{fig:IDS}, the framework of modern Chinese characters can be articulated through Ideographic Description Sequence (IDS)\footnote{\href{https://github.com/cjkvi/cjkvi-ids}{https://github.com/cjkvi/cjkvi-ids}}, which highlight 12 common structural motifs, such as left-right, top-bottom, and left-middle-right. These motifs provide insights into the spatial organization of radicals within a character. Armed with the deconstructed radicals from ancient texts, we apply the corresponding IDS to reassemble these radicals into the organized structure of modern Chinese characters, thus fulfilling the deciphering objective.

\begin{figure}[h]
    \centering
    \includegraphics[width=0.9\linewidth]{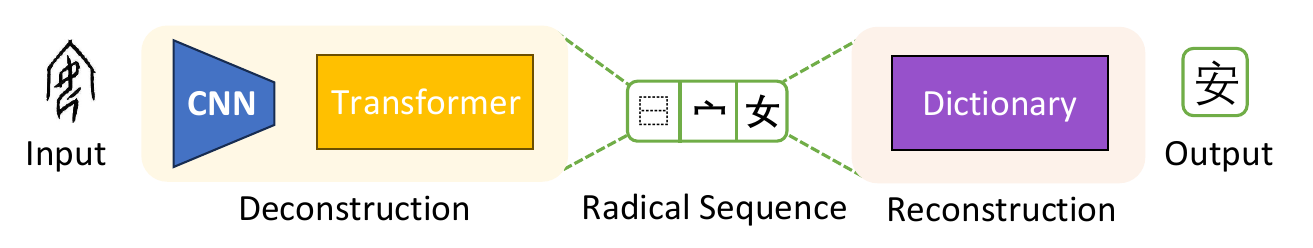}
    \vspace{-0.2cm}
    \caption{Overview of the network architecture of the proposed methods.}
    \label{fig:model}
    \vspace{-0.3cm}
\end{figure}
\vspace{-0.1cm}
\subsubsection{Network Architecture.} As depicted in Figure~\ref{fig:model}, the network architecture of our method is elegantly simple. Drawing inspirations from the scene text spotter method SPTS~\cite{peng2022spts,liu2023spts}, we have eliminated the coordination information typically required in text spotting tasks and have instead integrated IDS to form the predictive sequence. This means that for any given input image of an ancient Chinese character, the model is designed to output the deconstructed radical strokes along with the corresponding IDS. These elements are then utilized to reconstruct the modern standard Chinese characters, thus achieving the desired decipherment. To elaborate, given an input image $X \in {\mathbb{R}^{3 \times {{\rm{H}}_0} \times {W_0}}}$, the Deconstruction first employs a CNN to extract image features, producing a feature representation of $X^{1024\times 8\times 8}$. Subsequently, a Transformer module processes these features to deconstruct the ancient script into radical strokes and the possible IDS. Following this decomposition, the Reconstruction reassembles the identified radicals into modern Chinese characters. This reassembly is guided by predetermined IDS frameworks, ensuring each radical sequence is accurately matched to the correct character. This streamlined approach not only accelerates the reconstruction process but also enhances the accuracy of deciphering ancient scripts, making efficient use of the structural nuances of Chinese characters.

\vspace{-0.25cm}
\section{Experiments}
\vspace{-0.1cm}
\subsection{Implementation Details}

In our implementation, the ResNet-50 serves as the foundational backbone network. For the Transformer components, both the encoder and decoder are structured with six layers and feature eight attention heads, utilizing the Pre-Layer Normalization Transformer configuration as outlined by Xiong et al.~\cite{xiong2020layer}. Optimization is achieved through the AdamW optimizer, starting with an initial learning rate of 5$e^{-4}$ which is linearly decreased to a minimum of 1$e^{-6}$. The model is trained with a batch size of 256 for 100 epochs. Our demo will be available at \url{https://github.com/Pengjie-W/Puzzle-Pieces-Picker}

\vspace{-0.1cm}
\subsection{Deciphering Experiments}
\label{sec:Deciphering}

Given the intricate nature of deciphering ancient scripts, which often demands extensive validation by experts and presents challenges in quantitatively assessing the performance of proposed models, we opted for an alternative evaluation approach. We leveraged already deciphered ancient Chinese characters as test data to quantitatively analyze the performance of our model. Specifically, we assessed the deciphering capabilities of our proposed P$^3$ model across characters from seven distinct periods. To this end, we adopted a methodology similar to cross-validation, training seven different models for deciphering ancient Chinese characters, each tailored to a specific historical period.
\vspace{-0.4cm}

\begin{table}[ht!]
\centering
\begin{tabular}{l|c|c|c|c|c|c|c|c}
\hline
& OBIs&OBIs*    & Bronze  & Warring & Seal    & Clerical & Kangxi  & Regular \\ \hline
Sample-wise Acc.& 17.5\%&19.2\%  & 25.6\%  & 14.1\%  & 29.4\%  & 49.7\%   & 96.4\%  & 93.6\%  \\ \hline
\#Undeciphered        & 1,000&1000   & 682     & 592     & 935     & 491      & 848     & 999     \\
\#Success             & 394 &414    & 246     & 197     & 361     & 278      & 814     & 963     \\
Category-wise Acc.    & 39.4\%&41.4\%  & 36.1\%  & 33.3\%  & 38.6\%  & 56.6\%   & 96.0\%  & 96.4\%  \\ \hline
\end{tabular}
\caption{Quantitative results of the proposed P$^3$ on ancient character decipherment task. ‘OBIs*’ represents the experiments of deciphering OBIs after incorporating Radical Decomposition}
\label{tab:Deciphering}
\vspace{-0.8cm}
\end{table}

In detail, we commenced by randomly selecting 1,000 character categories from the dataset with the smallest sample size, referred to the OBI period, and designated these as \textit{undeciphered characters}. We then sought corresponding character classes in other historical periods to serve as the \textit{undeciphered characters} for those periods. However, due to the potential non-existence of certain characters across different eras, the number of \textit{undeciphered characters} in other periods might be less than 1,000. The total number of categories deemed as \textit{undeciphered} in each era is summarized in Table~\ref{tab:Deciphering}, illustrating variations across periods, with Bronze inscriptions featuring 682 classes while Seal script encompassing 935 classes. Subsequently, for each period's deciphering model, we formulated the training datasets by integrating two distinct components, while excluding the aforementioned categories of undeciphered characters. The first component comprised all other \textit{deciphered} characters from the same period within the ACCP dataset, aimed at aiding the model in learning the typographic features of that era. The second component consisted of data from all other periods, intended to facilitate the model's understanding of the evolutionary patterns of characters across different epochs. This setup led to the training of a total of seven distinct models, each evaluated for its effectiveness in deciphering characters from the respective periods. Furthermore, for the model of OBIs deciphering, due to its fewer categories, we conducted Radical Decomposition on the deciphered samples of 781 classes of OBIs, generating 9,787 radical samples, to further experiment with deciphering OBIs*. To evaluate the performance, we employed two primary metrics: sample-wise accuracy and category-wise accuracy. Sample-wise accuracy is defined as the ratio of correctly deciphered samples to the total number of samples, providing a granular view of the model's deciphering capability on an individual character basis. The latter metric deems a category successfully deciphered if any single sample within it is correctly interpreted, reflecting the practical value of the real-world decipherment process.

The experimental results, as presented in Table~\ref{tab:Deciphering}, highlight the varying degrees of deciphering success across different historical periods. Notably, the Kangxi Period exhibits the highest sample-wise accuracy at 96.4\% and a similarly high category-wise accuracy of 96.0\%. This outstanding performance can likely be attributed to the closer proximity of the Kangxi period to modern times, which entails a more consistent writing standard and an abundance of available data. In contrast, the OBIs period shows a very low sample-wise accuracy at 17.5\% and a category-wise accuracy of 39.4\%. After incorporating Radical Decomposition, the sample-wise accuracy increased to 19.2\%, and the category-wise accuracy increased to 41.4\%. This can be attributed to the significant age of the OBI period, which not only limits the availability of data but also introduces a greater intra-category variability. Characters from this era are more stylized and less standardized, posing a substantial challenge for deciphering efforts.

\begin{table}[t!]
\centering
\begin{tabular}{c|c|c|c|c|c|c|c|c}
\hline
OBI&Bronze& Warring&Seal&Clerical&Kangxi&Regular&OBIs Acc.&OBIs* Acc.\\ \hline
\checkmark & - & - & - &- & -&-&0.02\%&4.55\% \\ 
\checkmark & - & - & - &- & -&\checkmark&0.42\%& 4.88\%\\ 
\checkmark & - & - & - &- & \checkmark&\checkmark&0.45\%&4.54\% \\
\checkmark & - & - & - &\checkmark & \checkmark&\checkmark&0.80\%&4.96\% \\
\checkmark & - & - & \checkmark &\checkmark & \checkmark&\checkmark&2.53\%&6.64\% \\
\checkmark & - & \checkmark & \checkmark &\checkmark & \checkmark&\checkmark&6.23\%&9.24\% \\
\checkmark & \checkmark & \checkmark & \checkmark &\checkmark & \checkmark&\checkmark&17.47\%&19.16\% \\
\hline
\end{tabular}
\caption{Ablation study results showing the impact of including data from various historical periods on the decipherment accuracy of the OBI decipherment model. The ‘\checkmark’ symbol represents the usage of data corresponding to the respective period in the experiment, while the symbol "-" represents non-usage.}
\label{tab:ablation-obi}
\vspace{-0.8cm}
\end{table}

\vspace{-0.3cm}
\subsection{Ablation Study}

To further explore the efficacy of our model, particularly its capacity to learn the evolutionary patterns of ancient Chinese characters across different epochs, we selected the OBI decipherment model (including OBIs and OBIs*) as a case study for an in-depth ablation analysis.

\begin{table}[ht!]
\centering
\begin{tabular}{c|c|c|c|c|c|c}
\hline
OBIs*&Bronze& Warring&Seal&Clerical&Kangxi&Regular\\ \hline
4.78\% & 2.62\% & 0.96\% & 14.16\% & 10.91\% & 67.43\%&76.76\%\\ 
4.71\% & 2.95\% & 1.66\% & 16.25\% & 17.87\% & 92.63\%&\checkmark\\ 
4.75\% & 2.67\% & 1.48\% & 16.51\% & 21.01\% & \checkmark&\checkmark\\ 
4.97\% & 3.65\% & 2.31\% & 19.26\% & \checkmark & \checkmark&\checkmark\\ 
5.59\% & 5.47\% & 2.93\% & \checkmark & \checkmark & \checkmark&\checkmark\\ 
7.70\% & 15.06\% & \checkmark & \checkmark & \checkmark & \checkmark&\checkmark\\
19.16\% &\checkmark & \checkmark & \checkmark & \checkmark & \checkmark&\checkmark\\
\hline
\end{tabular}
\caption{Results of the extended ablation experiments. Each row represents one deciphering experiment, and the symbol ‘\checkmark’ indicates whether samples from the corresponding period were included in the training set.}
\label{tab:Ablation}
\vspace{-0.9cm}
\end{table}

In our ablation analysis presented in Table~\ref{tab:ablation-obi}, we meticulously examined the impact of systematically reintroducing samples from various historical periods into the training set of our OBI decipherment model. For the OBI model, starting with an initial setup that exclusively used OBI data, we observed a decipherment accuracy close to 0, highlighting the inherent difficulty of deciphering ancient scripts without a broader historical context.

Upon sequentially adding data from more recent periods, specifically the Regular and Kangxi scripts, we noted only a marginal improvement in accuracy. This modest increase, with accuracy reaching 0.42\% upon the inclusion of Regular script data and slightly fluctuating with the addition of Kangxi script data, suggests that the considerable temporal distance between these scripts and the OBI period limits their effectiveness in providing useful contextual information for decipherment. The narrative changed significantly with the inclusion of Clerical, Seal, and Warring States period data, where we observed a steady increase in accuracy, culminating at 6.23\%. This increment attests to the model's capability to extract and learn potential evolutionary patterns from cross-era data, thereby enhancing its decipherment performance.

The most pronounced surge in performance was witnessed with the incorporation of Bronze script data, which led to an 11.24\% jump in accuracy, bringing it to a notable 17.47\%. This substantial improvement underscores a strong correlation between the OBI and Bronze scripts, hinting at a direct lineage or significant shared characteristics that facilitate decipherment.

For the OBIs* model, as we sequentially introduced training data by more recent periods, the trend in decipherment accuracy was similar to that of the OBIs model. When data from the same period were added, the accuracy increased by approximately 2\% to 4\% compared to the OBIs model. This indicates that our method of decomposing components of Oracle Bone Inscription samples is effective.

Building upon the above insights, we further extended our investigation to a broader range of historical scripts, as detailed in Table~\ref{tab:Ablation}. This comprehensive examination reaffirmed the efficacy of our model in deciphering ancient Chinese characters by leveraging data from multiple eras. The consistent improvement in decipherment performance across various settings underscored the model's capacity to extract and learn from the evolutionary patterns of Chinese characters. This finding not only validates our approach but also highlights the importance of incorporating a diverse historical corpus to enhance the decipherment of ancient scripts.
\vspace{-0.2cm}
\subsection{Visualization}

\begin{figure}[t!]
    \centering
    \includegraphics[width=0.7\linewidth]{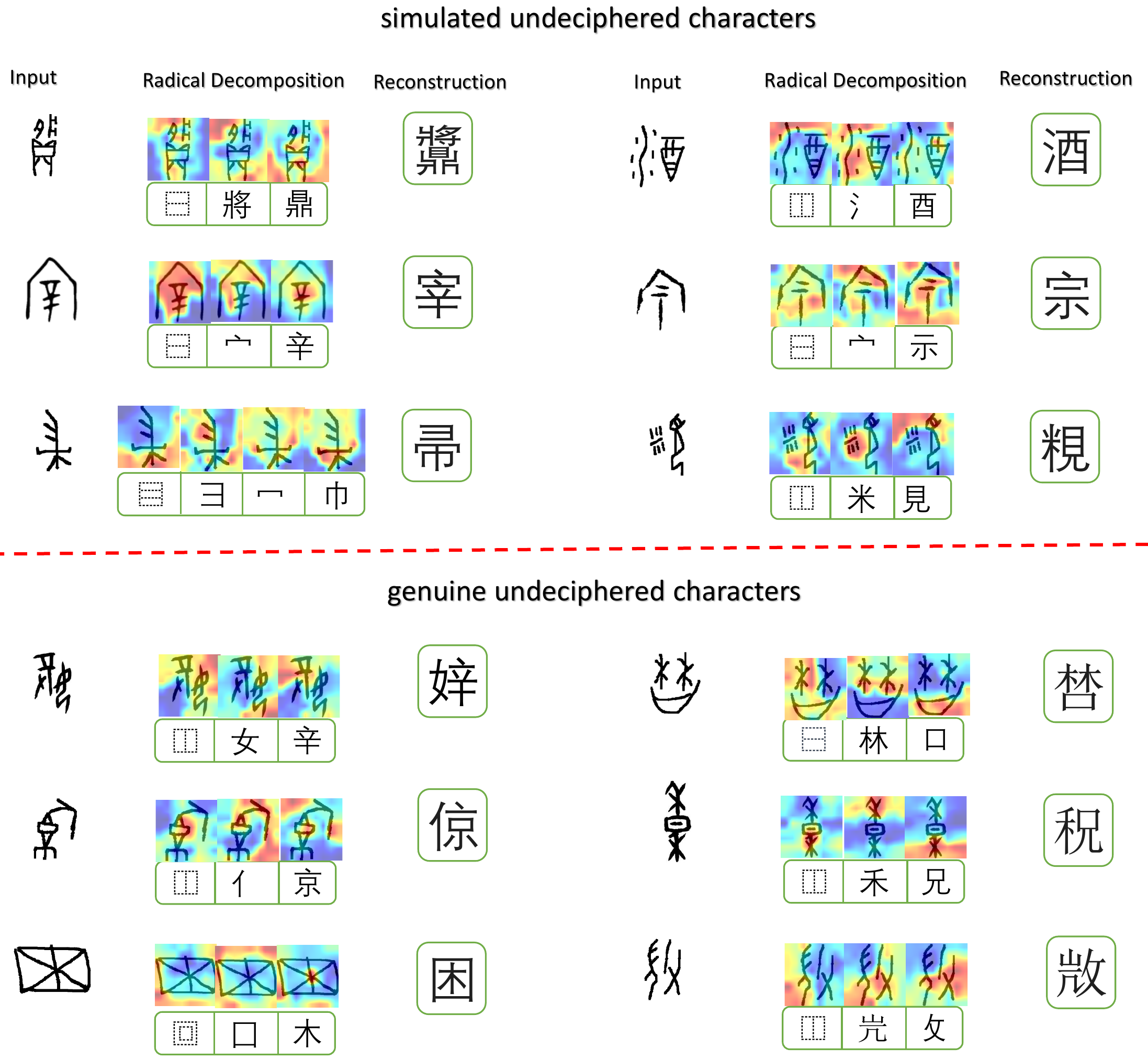}
    \caption{Inputting the oracle bone inscription images into the model yields the radical sequence as output, along with the corresponding visualization images for each component. In the visualization images, the intensity of red indicates the areas where the component is more relevant. 
In the upper half of the image, \textit{simulated undeciphered characters} represent images from a test set containing 1000 classes of OBI that have not been seen before. In the lower half, \textit{genuine undeciphered characters} represent truly undeciphered OBI images from the HUST-OBC dataset.}
    \label{fig:Visualization}
    \vspace{-0.5cm}
\end{figure}

To gain a more profound insight into our methodology, we employed Grad-CAM~\cite{selvaraju2017grad} for the visualization of the radical deconstruction and reconstruction process. As illustrated in Figure~\ref{fig:Visualization}, we demonstrate how the model deciphers unseen categories of OBI images. The model discerns radicals in a fashion similar to human intuition while also considering the proximity of components to predict their structural arrangement.
\vspace{-0.25cm}
\section{Conclusion}
\vspace{-0.2cm}
In this paper, we have introduced the Puzzle Pieces Picker, an innovative method inspired by jigsaw puzzles for deciphering ancient Chinese characters. This approach, leveraging the deconstruction and reassembly of radicals, was supported by the expansion of existing datasets to form a new ACCP dataset. Our tests on the ACCP dataset demonstrated the P$^3$ model's promising accuracy across different historical periods. Furthermore, our ablation study confirmed the model's effectiveness in identifying evolutionary patterns in ancient characters. While our focus was on ancient Chinese characters, the versatility of the P$^3$ model suggests its applicability to other radical-based writing systems, offering new avenues for linguistic research and historical document decipherment.
\vspace{-0.25cm}
\section*{Acknowledgments}
\vspace{-0.3cm}
This work was supported by the National Natural Science Foundation of China (No.62225603, No.62206104). This research is supported in part by National Natural Science Foundation of China (Grant No.: 62441604).

\end{document}